# Real-time Hyper-Dimensional Reconfiguration at the Edge using Hardware Accelerators

Indhumathi Kandaswamy*, Saurabh Farkya*, Zachary Daniels, Gooitzen van der Wal,
Aswin Raghavan, Yuzheng Zhang, Jun Hu, Michael Lomnitz, Michael Isnardi,
David Zhang, Michael Piacentino
Center for Vision Technologies, SRI International, Princeton, NJ
{Firstname.Lastname}@sri.com

## Abstract

*In this paper we present **Hy**per-**D**imensional **R**econfigurable **A**nalytics at the **T**actical **E**dge (**HyDRATE**) using low-SWaP embedded hardware that can perform real-time reconfiguration at the edge leveraging non-MAC (free of floating-point Multiply-ACcumulate operations) deep neural nets (DNN) combined with hyperdimensional (HD) computing accelerators. We describe the algorithm, trained quantized model generation, and simulated performance of a feature extractor free of multiply-accumulates feeding a hyperdimensional logic-based classifier. Then we show how performance increases with the number of hyperdimensions. We describe the realized low-SWaP FPGA hardware and embedded software system compared to traditional DNNs and detail the implemented hardware accelerators. We discuss the measured system latency and power, noise robustness due to use of learnable quantization and HD computing, actual versus simulated system performance for a video activity classification task and demonstration of reconfiguration on this same dataset. We show that reconfigurability in the field is achieved by retraining only the feed-forward HD classifier without gradient descent backpropagation (gradient-free), using few-shot learning of new classes at the edge.*

*Initial work performed used LRCN DNN and is currently extended to use Two-stream DNN with improved performance.*

## 1. Introduction

State-of-the-art (SOA) neural networks are incredibly complex, involving hundreds of layers with millions of hyperparameters. In edge-computing use cases, devices must operate under low-SWaP (Size, Weight, Power) constraints. Current large state-of-art deep neural networks (DNNs) do not scale to edge-computing use cases because they are both power hungry due to the floating point multiply-accumulate (MAC) operations and memory intensive due to needing to store millions of tunable network parameters [12]. In addition, deep neural networks are ill-suited for edge applications where the distributions of inputs and outputs can shift over time. Traditional neural networks are trained once (requiring long training times and using large amounts of high-compute resources) and deployed. Once deployed, they are rarely updated, and updating the models often requires re-training from scratch or finetuning the neural network over entire training datasets, including the data the network was previously trained on. In this work, we present **Hy**per-**D**imensional **R**econfigurable **A**nalytics at the **T**actical **E**dge (**HyDRATE**), an architecture designed to overcome these issues, and we demonstrate this architecture on low SWaP embedded hardware for the problem of video classification. Our approach relies on a non-MAC DNN encoder in combination with a non-MAC classifier based on Hyper-Dimensional (HD) computing described by Kanerva in [9] and Imani et. al in [2]. The proposed architecture retains near baseline accuracy on benchmark datasets for action recognition while enabling efficient reconfigurability in the field without backpropagation to handle significant shifts in the data distribution (e.g., adding a new class to the classifier).

The algorithmic contributions of our approach consist of two parts: 1) improved latency and power consumption via quantization of an existing deep neural network architecture such that all weights are powers of two using BitNet [10] and replacing the MAC operations with shift-accumulate (SACC) operations and 2) the ability to

___
"This research was, in part, funded by the Defense Advanced Research Projects Agency (DARPA). The views and conclusions contained in this document are those of the authors and should not be interpreted as representing the official policies, either expressed or implied, of the DARPA."

* Equal contribution.

reconfigure the network classifier for novel classes based on hyperdimensional computing [2].

The overall HyDRATE architecture is a sequence-to-sequence (seq2seq) model as shown in Figure 1. The left side is the non-MAC encoder; the right side is the HD classifier that acts as HD encoder during reconfiguration and HD decoder during inference. The network performs image- or video-based activity classification. The input is a sequence of video frames. The shallow non-MAC Sparse Spatial Encoder compresses an image into a set of feature vectors which, are filtered by a non-MAC temporal Neural Network resulting in a set of refined feature vectors. These temporal feature vectors are converted to HD vectors through XOR and binary logic operations. The output of the HD mapping in the decoder is compared with a set of class exemplars, hyper-vectors obtained in the training stage in the HD. The architecture is designed so that fast, forward-only (gradient-free) training of the HD encoder can be performed for recognition of new activity classes.

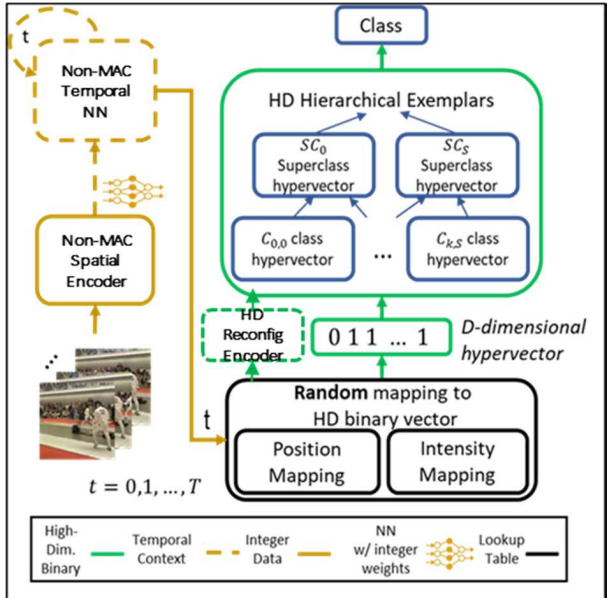

Figure 1 HyDRATE Architecture

- Within our architecture all data in Neural Net is 8-bit integer type with non-MAC operations
- The dataflow is like a standard seq2seq model
- The non-MAC Sparse Encoder like ResNet50 DNN, converts an image frame into spatial feature vectors
- The non-MAC Temporal NN like LSTM or optical flow ResNet50 produces features with temporal extent over time 't' frames
- Feature vectors are converted to HD vectors through XOR and binary bundle operations
- The output of the HD mapping is compared with a set of class exemplars, hyper-vectors obtained in the training stage in the HD classifier. We use Hamming distance as the comparison metric
- New class exemplars are created on demand by HD Reconfiguration Encoder without back propagation

The steps to create the non-MAC DNN and training of the HD class exemplars are explained by Isnardi et al. in [1]. In this paper, we will discuss briefly on the algorithmic concepts and network performance with details on the hardware implementation and the measured performance. The UCF101 dataset and videos captured using mobile phones and webcam representing the activity were used for training, inference, reconfiguration, and evaluation of the hardware.

To the best our knowledge we did not find a similar hardware software study on non-MAC architectures for activity recognition. Parajuli et. al [3] has comparison to other quantized architectures for classification.

## 2. Network quantization, inference, and reconfiguration

### 2.1. Training a non-MAC quantized DNN

As mentioned earlier, the model consists of two parts: 1) a DNN-based feature encoder with low-precision weights and Shift-ACCumulate (SACC) operations in place of MAC operations and 2) an HD classifier for performing the activity classification task. In this section, we discuss the details of the non-MAC feature extractor. In order to use SACC operations within the DNN, the network weights must be Powers-of-Two (PoT) instead of floating-point values. Thus, we must learn to quantize the network, where network quantization [11] involves compressing the high-precision weights of a neural network while maximizing the low-precision/compressed network's accuracy.

We create a non-MAC SACC DNN by implementing a distilled teacher-student network as described by Parajuli et. al in [3]. The teacher NN is a 32-bit floating-point network, and the student NN is a quantized low-precision network with PoT coefficients. To obtain PoT coefficient weights, we use a learnable quantization introduced by Parajuli et. al in [3]. Further, we use linear quantization in activation layers, restricting the activations to 8-bits.

This architecture learns both student and teacher models simultaneously from scratch. Although, our strategy is applicable to any DNN, in our use case, we perform activity recognition using LRCN described by Donahue et. al in [4] on the UCF101 dataset described by Soomro et al. in [6]. ResNet50, as explained by He et. al in [5] and as shown in Figure 2 extracts visual features followed by LSTM for sequence learning.

Once the training is done, we freeze the student network to obtain a non-MAC DNN in PoT. During inference, the quantized ResNet50 extracts spatial features per frame, and the quantized LSTM extracts temporal information from the spatial features.

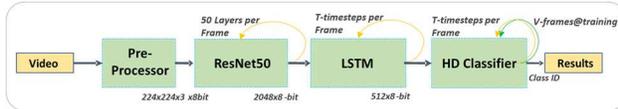

Figure 2 LRCN Encoder + HD Classifier

We are currently working on training and implementing an improved two-stream ConvNet described by Simonyan et al. in [7]. Instead of using a ConvNet in combination with an LSTM for combining spatial feature and temporal feature extraction, this network uses two ConvNets in parallel. As shown in Figure 3, one branch is a standard ResNet50 model operating over RGB image frames to capture spatial information, and the second branch is a ResNet50 model that operates over the optical flow maps of the corresponding image frames to extract temporal information. The model then performs late fusion to predict the activity class of a video.

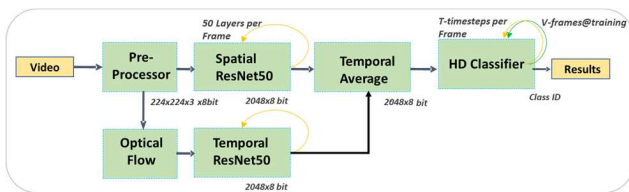

Figure 3 Two-stream Encoder + HD Classifier

Feature tracking using optical flow is described by Jean-Yves Bouguet in [8]. The two-stream network is superior in performance compared to LRCN network because two-stream network explicitly encodes the optical flow information from the videos while LRCN does this in feature space using an LSTM.

### 2.2. Training the HD classifier and inference

A detailed method of LRCN and HD training, and inference formulation is explained by Isnardi et al. in [1]. Briefly, HyDRATE's HD backend was inspired by the VoiceHD architecture explained by Imani et al. in [2]. While VoiceHD encodes voice signal to a HD vector in frequency domain, our architecture encodes the values and positions of each feature of the feature vector (per frame) out of the non-MAC NN to a HD vector (per frame) in time domain.

For the UCF101 dataset, 101 HD exemplars are generated during training, one exemplar per class. To train the HD model, we use majority voting to bundle all HD vectors of all the samples of a class from the training set to form a class HD exemplar. For video activity recognition, HyDRATE combines the HD vectors of a sliding 12-frames temporal window and averages the Hamming distance output to classify the activity. Note that HD training is performed using feedforward bitwise logical operations, and traditional backpropagation is not used.

A key innovation in our proposed architecture lies in increasing the richness of operations performed with the HD representation. Previous approaches have used HD computations in a limited capacity, mainly as a way of performing a nearest-neighbor search largely defined by the non-HD input features. In contrast to these works, our use of HD computing enables reconfigurability of the classifier.

### 2.3. Reconfiguration

In this section, we introduce the new concept of reconfiguration in HD. The idea is to use the pre-trained non-MAC model as the feature extractor and perform on-demand training on the hardware to create a new set of HD class exemplars, assuming the input data distribution of the new classes are like that of training classes. For example, if we train a model on 99-classes using UCF101 dataset we can perform reconfiguration on the remaining two classes using the same dataset or add new data of an existing class from a different dataset. The intuition is that the new data or class for online training will have similar low-level and mid-level features as that of training class and the HD classifier can exploit those features and create a discriminative classifier. Performing this kind of online training is much faster and efficient in the HD domain as shown in the results in Figure 4 with K-shot learning, on a video activity.

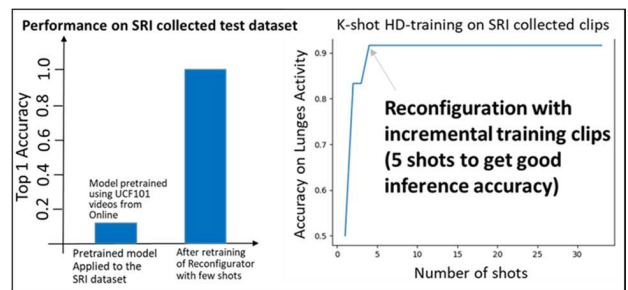

Figure 4 K-shot learning

## 3. Network performance

### 3.1. Algorithm experimental setup

We use the UCF101 official test setup 1 to evaluate our approach. This data split consists of a total of 9518 training videos and 3769 test videos for 101 classes. Each video clip has a frame rate of 25fps. In our experiments, we sample the videos such that we have 50 frames per clip fixed for training and testing. We train our models with fixed learning rate of 1e-3 and use ImageNet pretrained weights for ResNet50 to do transfer learning. All our models are trained for 15-epochs and the best model is selected based on the lowest validation loss between non-MAC DNN and Ground Truth.

## 3.2. Non-MAC DNN performance

We first evaluate the trade-offs between accuracy and precision when quantizing the LRCN model. Figure 5 shows the performance comparison of the 32-bit floating point baseline model (blue line) and quantized models (Orange line). The orange line shows the performance of the non-MAC DNN with increasing value of $\lambda_{bit\_loss}$ for weights. Using the student-teacher loss formulation explained by Parajuli et. al in [3], as we increase the value of $\lambda_{bit\_loss}$ (hyper param to control bit / quantization loss) we inject more quantization in the model resulting in the lower number of bits as shown on the x-axis. Further, we fixed $\lambda_d$ = 0.9 (hyper parameter for distillation loss).

The quantized model with an average 2.79-bits ($\lambda_{bit\_loss}$ =2e-4) was chosen to be used in the FPGA hardware, the reason being this model was targeted to quantize to 4-bits or lower per layer. Interestingly, some of the weight matrices are just 2-bits.

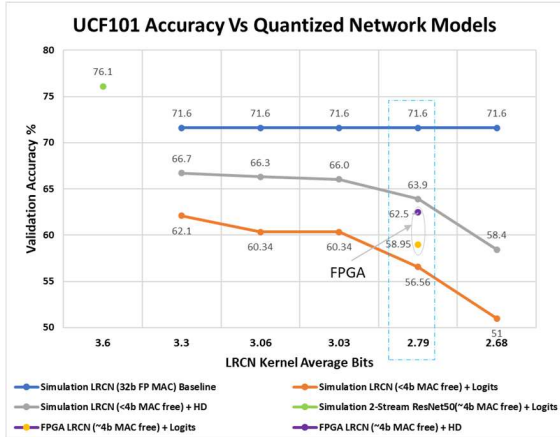

Figure 5 Accuracy vs Network Model Size

## 3.3. HD performance

Next, we look at the effect of combining the quantized LRCN (and 2-stream ResNet50) in combination with the HD classifier vs using a standard set of fully connected layers for classification at the end of the network. For HD training, K features per frame from the non-MAC DNN was used to generate a 'D' dimensional HD vector. HD vectors from the F-consecutive frames within the sliding window were used to classify the activity. In LRCN+HD network, K=512 features, F=12 frames. In 2-Stream ResNet50+HD network, K=256 features, F=10 frames. There is an MLP layer at the output of 2-Stream ResNet50 that converts the 2048-features to 256-features. Figure 6 shows the accuracy of the networks for UCF101 video activity classification with varied HD exemplar length 'D' from 256-bits to 4096-bits for encoding the features in HD classifier. From the results, we saw the performance saturates after the 4096-bit HD vector in network (Figure 6, orange line). We saw an improved accuracy over 32-bit floating point MAC LRCN baseline network (Figure 6, blue line) using quantized 2-stream ResNet50 +logits (Figure 6, purple line).

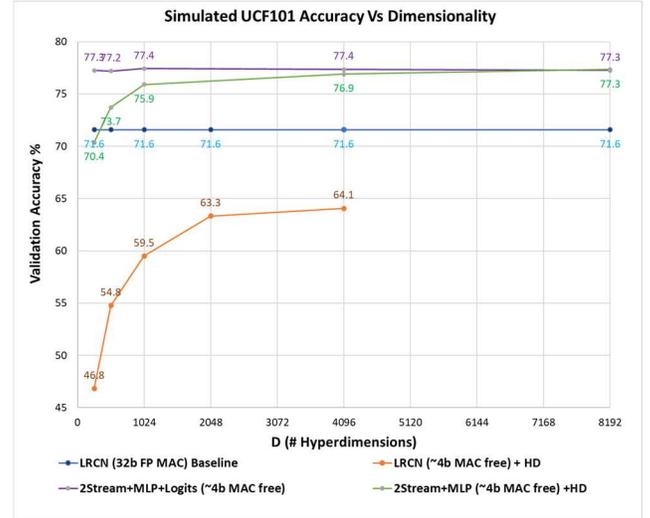

Figure 6 Simulated Accuracy vs Dimensionality LRCN, 2-Stream ResNet50 & HD network for UCF101 video activity classification

Interestingly, for a chosen network model, using the HD classifier also boosts the performance of model from 56.6% (Figure 5, orange line) to 63.9% (Figure 5, grey line) in LRCN network. Note that the 2-Stream ResNet50+HD (Figure 6, green line) overtakes 2-Stream ResNet50+logits (Figure 6, purple line) for HD vector length 8192-bits. This indicates the HD's ability to discriminate in the high dimensional binary space even with noise in the feature space.

Figure 7 shows the performance accuracy after reconfiguration of the 100th UCF101 class using the 2-Stream ResNet50+HD network using K=256 features, D=4096-bits and F=10 frames. It can be seen there is a 0.4% drop in the overall accuracy. We have noticed that after adding new class, the pretrained model may suffer a slight performance degradation on the old classes, since sample distribution of new classes may overlap with the old classes. An easy solution to boost the reconfiguration performance can be done by retraining the HD class exemplars when there are wrong matches [2]. This needs to be done in one epoch.

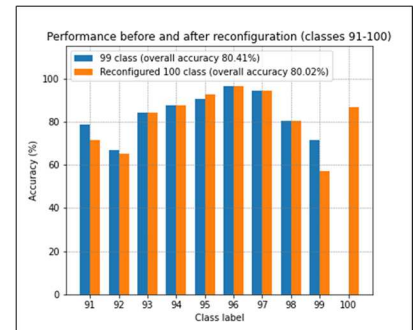

Figure 7 Results of Reconfiguration of UCF101 100th class using 2-Stream ResNet50 & HD Network

## 4. Hardware Implementation

### 4.1. Hardware framework

To demonstrate HyDRATE, we chose iWaveSystems's G35D FPGA development platform with a Xilinx ZU19EG Zynq UltraScale+ FPGA, as shown in Figure 8, that has a combination of embedded processors and programmable logic. Figure 9 shows the mapping of the LSTN + HD processing on this platform. The non-MAC ResNet50, LSTM and HD classifier are implemented as hardware accelerators. The ResNet50 accelerator runs per layer computations while the LSTM and HD classifier run per timestep computations. These accelerators interface to the external memory through DMA modules and are controlled by the Zynq's low-latency real-time processors. The video interface to the board is through a USB camera. The demo scripts are executed through a serial command line interface. The real-time processor in Zynq, running light-weight OS, controls the accelerators, movement of video, and model data. At any point in time, the user can both initiate and stop reconfiguration to create a new video activity class.

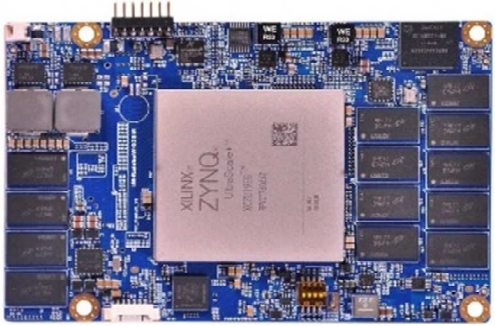

Figure 8 FPGA Demonstration Platform

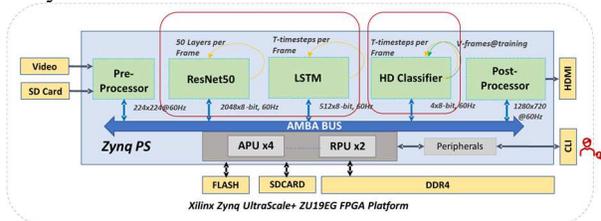

Figure 9 Framework using LRCN with HD Classifier

We are in the process of updating this configuration for the Two-Stream Convnet & HD classifier, as shown in Figure 10, using the same hardware accelerator components.

There are existing frameworks for FPGA accelerators in DNN [13],[14],[15],[16], LSTM [17],[18],[19] and HD computing [20],[21]. The novelty with our work is 1) NNPE/LSTM accelerator performs single SACC operation per weight of the kernel on 8-bit data, scalable SACC vectors maintains full internal precision; and reduced power due to SACC and near memory computation with minimal movement of layer data to external memory 2) HD encoder-decoder performs real-time reconfiguration with short latency 3) Accelerators have generic AXI and CPU interfaces.

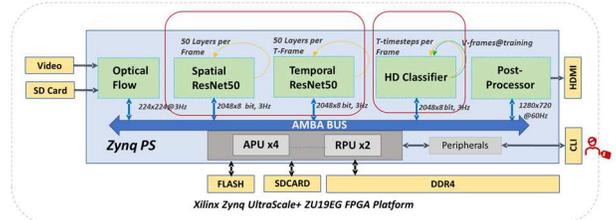

Figure 10 Framework using Two-stream ConvNet and HD Classifier

#### 4.1.1 Inference and Reconfiguration steps

In the hardware setup, the LRCN network was loaded with a 4-bit quantized network model that was trained for the first 99-classes using the UCF101 training dataset. For inference, the HD Classifier was loaded with pre-trained model that was trained for the three classes in UCF101 using UCF101 training data and mobile phone captured training videos. To test inference, the USB camera was focused on the monitor playing UCF101 test videos and mobile phone captured test videos. The classification result from the HD classifier was overlayed on the display.

For reconfiguration at the edge, pre-recorded training videos for the "writing-on-board" that represents 100$^{th}$ UCF101 class was used. These training videos were captured with mobile phones, showing reconfigurability across both novel classes and novel sensors. Inference performance before and after reconfiguration was tested using a mix of UCF101 test data and mobile phone captured test data.

#### 4.1.2 Neural Network Processing Engine (NNPE)

Both ResNet50 and LSTM modules are implemented using the configurable Neural Network Processing Engine (NNPE) that includes 'S' configurable vector modules based on shift-and-accumulate (SACC) architecture achieving 10x simpler operations than multiply-accumulate (MAC) functions. Figure 11 shows the diagram of the NNPE with a configurable data input buffer, parameter buffer for each SACC module, and configurable data output buffer. Each vector SACC module has N parallel sign, shift, and add functions, and an accumulator to perform M×N vector operations. The input buffer and parameter buffer are organized to provide M×N wide data representing k × k convolutions for Cin channels, where M × N ≥ k × k × Cin. All SACC modules operate on the same input data, generating S output channels. This operation is repeated Cout(Channels)/S times. The resulting data is stored in the output buffer, organized to provide the data for the next layer after swapping the input and the output buffers. In the implemented hardware, NNPE was configured with S=8 processing arrays and N=256 SACC computations per array.

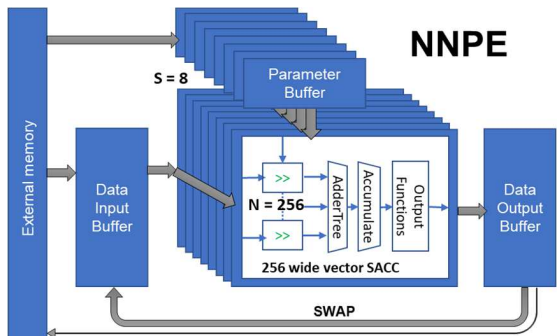

Figure 11 Implemented NN Accelerator Configuration

For most of the layer operations, the data can remain in the swappable embedded data buffers. Only the first input data, the final output data, and some of the intermediate data are read and/or written to external memory, which saves a significant amount of power. By adding a third Data buffer saving intermediate layer data (as is required in ResNet50), data access to external memory can be further reduced, saving additional power and latency.

*4.1.3   LSTM and other networks*

The vector SACC module can be programmed and configured for any of the ResNet50 layers or for many of the common CNN type layers. The output functions contain options to compute a Batch Normalization function, scaling of the data, a RELU operation, and non-linear operators (such as sigmoid and tanh functions) that are required by LSTM or other functions. For this FPGA implementation, the NNPE is separately configured for ResNet50 and a parallel operating LSTM function. NNPE could also be implemented as a software configurable module in an ASIC for a wide set of applications. For this implementation, the LSTM accelerator is a modified NNPE version for 160 SACC vector computations. Sigmoid and tanh functions are implemented using LUT approximations.

*4.1.4   ASIC implementation*

This architecture lends itself well for an ASIC implementation, which can be scaled by selecting the number of parallel NNPE modules, including a rich set of output functions, and would require only a very small real time controller for programming the processing flow. The data buffers can be implemented with highly optimized memory components, or the SACC with memory components could be implemented as a PIM (Processor in Memory) architecture in a future ASIC. An ASIC implementation would provide significant power savings over the FPGA implementation (estimated at >5x using the same technology node) and can process at >3x clock rate.

### 4.2. Software framework

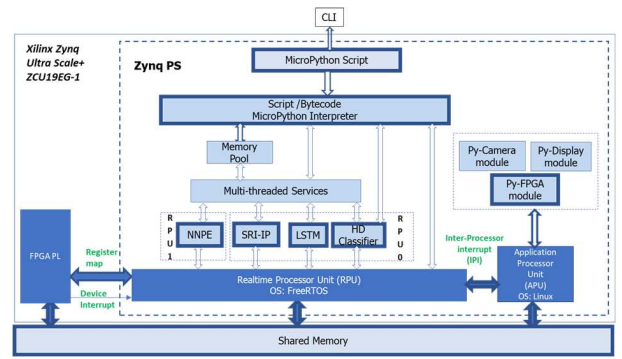

Figure 12 Software Framework

The software framework for the FPGA implementation is shown in Figure 12. The APUs run Linux and RPUs run FreeRTOS. The processors interact through IPI communication, while the accelerators communicate through device interrupts or through polling device registers. There is a shared external memory between the FPGA accelerators and the processors. MicroPython scripts are used for system and user configuration. Device drivers are encapsulated in python wrappers to be used as python modules. The Zynq processors can be replaced with MICO32 processors for small footprint FPGA and to get further reduction in system latency.

### 4.3. Memory bandwidth

Table 1 shows the memory size for the RESNET50 implementation, and compared the total data size for image, layer, and parameter data per frame between SACC and 32-bit MAC implementation. The data in brackets [*] indicates the data size saved by using embedded data buffers in our NNPE architecture. The data bandwidth is >10x less for the SACC implementation, assuming the same frame rate, providing a significant external memory power reduction. The average and peak memory bandwidth can be accommodated by a single FPGA DDR4 4GB module. And could support a much lower power memory with a future ASIC.

| External memory access | SACC | MAC |
|---|---|---|
| Image from mem [Mbit] | 1.2 | 1.2 |
| Layers from mem [Mbit]* | 0 [75] | 300 |
| Layer to mem [Mbit]* | 0 [75] | 304 |
| Scratch to mem [Mbit] | 5 | 20 |
| Scratch from mem [Mbit] | 5 | 20 |
| Parameters from mem [Mbit] | 194 | 1526 |
| Total data from/to mem [Mbit]* | 205 [355] | 2171 |
| Data BW saving if frame rate is same | 10.6 [6.1] | 1 |

Table 1 DDR Bandwidth

## 5. Performance measurements

### 5.1. Latency measurements

With an FPGA processing clock at 187.5MHz, the processing time for the ResNet50 is 40.28ms per frame, LSTM is 3.017ms per timestep, and HD is 0.193ms per timestep. When eleven of the twelve time-steps required for the LSTM were computed in parallel with the ResNet50 computation, the total latency is less than 50 msec (20 Hz frame rate) in this FPGA implementation. The ResNet50 execution would reduce to less than 20 msec if a third internal buffer is added, as indicated in NNPE description above. The LSTM time can be made shorter by parallelizing the current and recurrent kernel loads from the external memory.

### 5.2. Power measurements

System power was measured using a COTS DC power meter. The measured power of the programmable logic running LRCN network using NNPE without external DDR access is 3.9W. With further optimization of the HD classifier, the estimated power of LRCN + HD network would be 3.1W. Extrapolating the logic to use MAC instead of SACC and estimating the power using the Xilinx Vivado power estimator tool, we would get a 5.9x power reduction and 4x latency reduction, resulting in a combined 24x power x latency reduction. Going to an ASIC will provide an additional >5x improvement [22] resulting in a >120x combined power and latency reduction.

### 5.3. Accuracy vs model size vs HD dimension

Performance of FPGA hardware loaded with a pre-trained 2.79-bit LRCN network model and a 4096-bit HD model was evaluated using UCF101 test dataset. Figure 5 shows that inference hardware accuracy of 62.5% to be close to the simulation accuracy of 63.9%, validating the implemented hardware system.

### 5.4. Noise measurements

Important metrics for this application are sensitivity to noise and HD vector bit-flipping. The graph in Figure 13 shows the accuracy results of the UCF101 test set with white gaussian noise added to the data set. The data shows that the design is noise tolerant down to 40dB SNR. The graph in Figure 14 shows the accuracy to bit-flipping of encoded HD vectors. The data indicates that the design is tolerant up to 30% bit flipping.

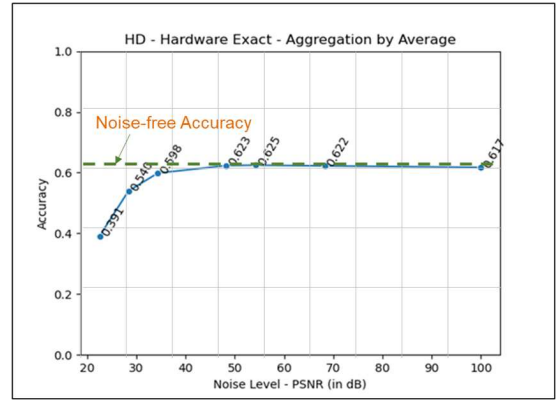

Figure 13 Accuracy vs White Noise

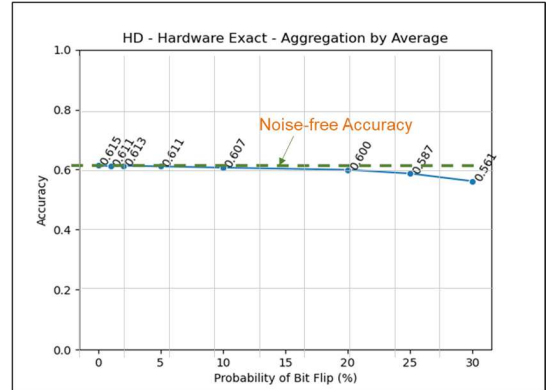

Figure 14 Accuracy vs HD Bit Flipping

### 5.5. Utilization

FPGA utilization of the demo system is shown in Table 2. Note that the HD classifier is very large and can be significantly reduced with minimal effect on latency.

| Zynq US+ ZCU19EG Utilization | LUT % | FF % | DSP48 % | BRAM % | URAM % |
|---|---|---|---|---|---|
| NNPE 8x, SACC256 | 20.79 | 12.30 | 1.32 | 14.18 | 50.00 |
| LSTM 1x, SACC160 | 1.43 | 1.01 | 0.15 | 3.96 | 0.00 |
| HD Classifier, 4096-bit HV | 21.43 | 11.60 | 0.00 | 17.43 | 0.00 |
| Pre-Processing | 0.15 | 0.24 | 0.00 | 0.00 | 0.00 |
| Zynq AXI devices | 7.03 | 5.81 | 0.00 | 6.91 | 0.00 |
| Total Utilization | 50.83 | 30.97 | 1.47 | 42.48 | 50.00 |

Table 2 Xilinx ZU19EG Utilization

## 5.6. Initial Two-stream ResNet50 and HD classifier implementation

Replacing the Resnet50 + LSTM with a two-stream ResNet50 approach, as indicated in Section 2.1, improves performance. The Spatial ResNet50 and the Temporal ResNet50 can use the same NNPE, with the input to the Temporal ResNet50 provided by a packed 10-frame motion vector from the Optical Flow computation module. The total ResNet50 operation would slow down by only 10%. For development purposes, optical flow and HD classifier are first implemented in a PC, while spatial and temporal ResNet50 are executed in the FPGA board using NNPE, as shown in Figure 15.

While the power is estimated to remain the same, compared to the LRCN network it takes a tenth more of time to compute the combined spatial and temporal feature vectors.

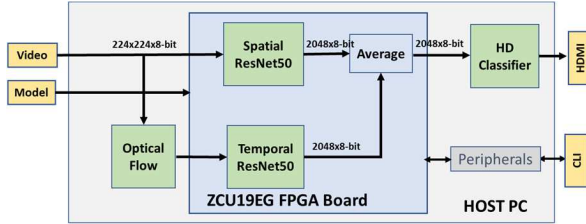

Figure 15 Two-stream ResNet50 + HD Classifier

## 6. From Algorithmic Concept to Real-Time Demo

Figure 16 shows the steps followed from algorithmic concept to the real-time demo. The first step was to train LRCN and HD network in the Tensorflow-Keras framework using the UCF101 dataset to get full precision LRCN and HD network models. The second step was to quantize the full-precision LRCN model to a PoT reduced bit model. The third step was to architect an optimized system considering the memory bandwidth, processing speed, latency, and power. The resulting design is implemented in embedded hardware and software using Xilinx Vivado environment and verified against bit-accurate simulation data generated using NumPy scripts. The final step was to integrate the hardware, embedded software and quantized NN model and test them on the targeted FPGA platform using a live camera feed.

## 7. Summary

HyDRATE architecture is a non-MAC feature extractor combined with a HD classifier, that can also perform reconfiguration at the edge without gradient descent. HyDRATE has been optimized for video classification, and its performance is on par or better than MAC-based DNNs such as ResNet50 + LSTM and 2-Stream ResNet50. The authors believe this is the first non-MAC HD architecture to perform 3D (video) analytics at video rates. Its non-MAC, logic-based FPGA implementation addresses the low-power needs of edge-friendly deployment while also providing edge-friendly retraining of the HD decoder.

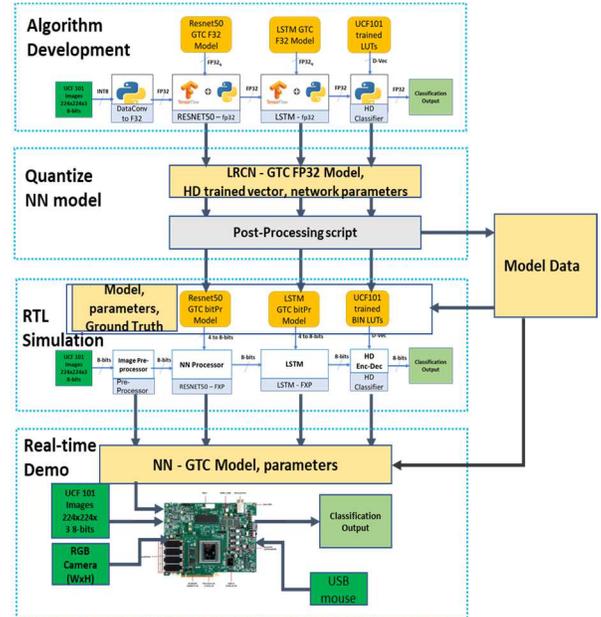

Figure 16 Steps from Concept to Real-Time Demo

## 8. Acknowledgment

We would like to thank DARPA for giving SRI the opportunity to explore and present new concepts in hyperdimensional computing as part of the DARPA's Artificial Intelligence Exploration (AIE) HyDDENN program.

## References

[1] Michael Isnardi et al., "Hyper-Dimensional Analytics of Video Action at the Tactical Edge". Presented at the *GOMACTech 2021 conference*, March 29 – April 1, 2021.
[2] M. Imani et al., "VoiceHD: Hyperdimensional Computing for Efficient Speech Recognition", *ICRC*, 2017.
[3] Parajuli, S., Raghavan, A. and Chai, S. "Generalized Ternary Connect: End-to-End Learning and Compression of Multiplication-Free Deep Neural Networks." arXiv preprint arXiv:1811.04985 (2018).
[4] Donahue, Jeffrey, et al. "Long-term recurrent convolutional networks for visual recognition and description." *Proceedings of the IEEE conference on computer vision and pattern recognition*. 2015.
[5] He, Kaiming, et al. "Deep residual learning for image recognition." *Proceedings of the IEEE conference on computer vision and pattern recognition*. 2016.
[6] Soomro, Khurram, Amir Roshan Zamir, and Mubarak Shah. "UCF101: A dataset of 101 human actions classes from videos in the wild." *arXiv preprint arXiv:1212.0402* (2012).

[7] Simonyan, Karen, and Andrew Zisserman. "Two-stream convolutional networks for action recognition in videos." *Advances in neural information processing systems* 27 (2014).

[8] Bouguet, Jean-Yves. "Pyramidal implementation of the affine Lucas Kanade feature tracker description of the algorithm." *Intel corporation* 5.1-10 (2001): 4.

[9] Kanerva, Pentti. "Hyperdimensional computing: An introduction to computing in distributed representation with high-dimensional random vectors." *Cognitive computation* 1.2 (2009): 139-159.

[10] Raghavan, Aswin, et al. "Bitnet: Bit-regularized deep neural networks." *arXiv preprint arXiv:1708.04788* (2017).

[11] Gholami, Amir, et al. "A Survey of Quantization Methods for Efficient Neural Network Inference." *Low-Power Computer Vision*. Chapman and Hall/CRC 291-326.

[12] Hadidi, Ramyad, et al. "Characterizing the deployment of deep neural networks on commercial edge devices." *2019 IEEE International Symposium on Workload Characterization (IISWC). IEEE, 2019*

[13] Zhang, Chen & Li, Peng & Sun, Guangyu & Guan, Yijin & Xiao, Bingjun & Cong, Jason. (2015). Optimizing FPGA-based Accelerator Design for Deep Convolutional Neural Networks. 161-170. 10.1145/2684746.2689060.

[14] Chang, Sung-En et al. "Mix and Match: A Novel FPGA-Centric Deep Neural Network Quantization Framework." *2021 IEEE International Symposium on High-Performance Computer Architecture (HPCA)* (2021): 208-220.

[15] Hao, C., Zhang, X., Li, Y., Huang, S., Xiong, J., Rupnow, K., Hwu, W.W., & Chen, D. (2019). FPGA/DNN Co-Design: An Efficient Design Methodology for 1oT Intelligence on the Edge. *2019 56th ACM/IEEE Design Automation Conference (DAC), 1-6.*

[16] R. DiCecco, G. Lacey, J. Vasiljevic, P. Chow, G. Taylor and S. Areibi, "Caffeinated FPGAs: FPGA framework For Convolutional Neural Networks," 2016 *International Conference on Field-Programmable Technology (FPT)*, 2016, pp. 265-268, doi: 10.1109/FPT.2016.7929549.

[17] Y. Guan, Z. Yuan, G. Sun and J. Cong, "FPGA-based accelerator for long short-term memory recurrent neural networks," *2017 22nd Asia and South Pacific Design Automation Conference (ASP-DAC),* 2017, pp. 629-634, doi: 10.1109/ASPDAC.2017.7858394.

[18] J. He, D. He, Y. Yang, J. Liu, J. Yang and S. Wang, "An LSTM Acceleration Engine for FPGAs Based on Caffe Framework," *2019 IEEE 5th International Conference on Computer and Communications (ICCC),* 2019, pp. 1286-1292, doi: 10.1109/ICCC47050.2019.9064358.

[19] *Chang, A.X., Martini, B., & Culurciello, E. (2015). Recurrent Neural Networks Hardware Implementation on FPGA. ArXiv, abs/1511.05552.*

[20] Salamat, S., Imani, M., & Rosing, T. (2020). Accelerating Hyperdimensional Computing on FPGAs by Exploiting Computational Reuse. *IEEE Transactions on Computers, 69, 1159-1171.*

[21] Imani, Mohsen et al. "FACH: FPGA-based acceleration of hyperdimensional computing by reducing computational complexity." *Proceedings of the 24th Asia and South Pacific Design Automation Conference (2019)*

[22] Kuon, I and Rose, J, "Measuring the Gap Between FPGAs and ASICs," *IEEE transactions on Computer-aided Design of IC and Systems, 2007*